\documentclass[10pt,twocolumn,letterpaper]{article}

\usepackage{iccv}
\usepackage{times}
\usepackage{epsfig}
\usepackage{graphicx}
\usepackage{amsmath}
\usepackage{amssymb}
\usepackage{booktabs}
\usepackage{caption}
\usepackage{subcaption}
\usepackage{booktabs}
\usepackage{siunitx}
\usepackage{multirow} 
\usepackage{color,soul} 

\usepackage[breaklinks=true,bookmarks=false]{hyperref}

\iccvfinalcopy 

\ificcvfinal\fi

\begin{document}

\title{MoVi: A Large Multipurpose Motion and Video Dataset}


\author{Saeed Ghorbani \\
York University \\
Toronto, ON, Canada \\
{\tt\small saeed@eecs.yorku.ca} 
\and
Kimia Mahdaviani \\
Queen's University \\
Kingston, ON, Canada \\
\and
Anne Thaler \\
York University \\
Toronto, ON, Canada \\
\and
Konrad Kording \\ 
University of Pennsylvania \\
Philadelphia, USA \\
\and
Douglas James Cook \\ 
Queen's University\\
Kingston, ON, Canada \\
\and
Gunnar Blohm \\
Queen's University \\
Kingston, ON, Canada \\
\and
Nikolaus F. Troje \\
York University \\
Toronto, ON, Canada \\
}

\maketitle
\ificcvfinal\thispagestyle{empty}\fi

\begin{abstract}
Human movements are both an area of intense study and the basis of many applications such as character animation. For many applications it is crucial to identify movements from videos or analyze datasets of movements. Here we introduce a new  human Motion and Video dataset MoVi, which we make available publicly. It contains 60 female and 30 male actors performing a collection of 20 predefined everyday actions and sports movements, and one self-chosen movement. In five capture rounds, the same actors and movements were recorded using different hardware systems, including an optical motion capture system, video cameras, and inertial measurement units (IMU). For some of the capture rounds the actors were recorded when wearing natural clothing, for the other rounds they wore minimal clothing. In total, our dataset contains 9 hours of motion capture data, 17 hours of video data from 4 different points of view (including one hand-held camera), and 6.6 hours of IMU data. In this paper, we describe how the dataset was collected and post-processed; We present state-of-the-art estimates of skeletal motions and full-body shape deformations associated with skeletal motion. We discuss  examples for potential studies this dataset could enable. 
\end{abstract}
\raggedbottom
\section*{Keywords}
Human motion dataset, optical motion capture, IMU, video capture
\begin{figure*}
  \centering
  \begin{subfigure}{\linewidth}
  \begin{subfigure}{0.25\textwidth}
    \centering
    \includegraphics[width=0.9\textwidth]{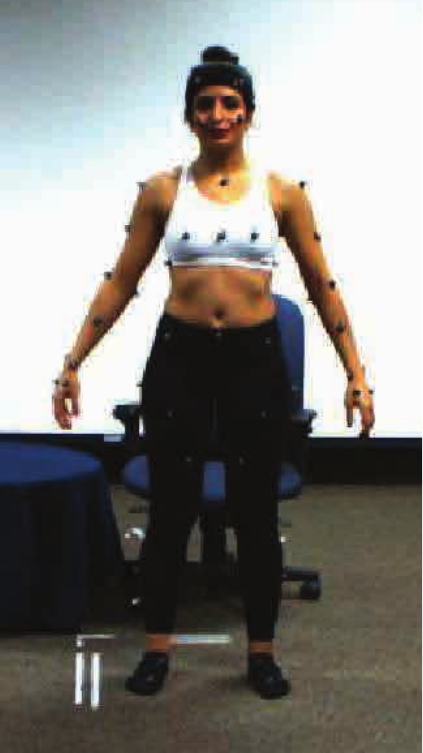}
\label{fig:sfig1}
  \end{subfigure}%
  \begin{subfigure}{0.25\textwidth}
    \centering
    \includegraphics[width=0.9\textwidth]{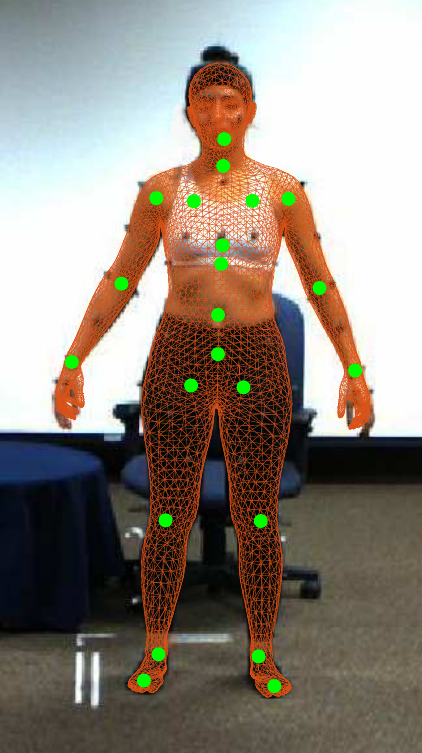}
    \label{fig:sfig1}
  \end{subfigure}%
  \begin{subfigure}{0.25\textwidth}
    \centering
    \includegraphics[width=0.9\textwidth]{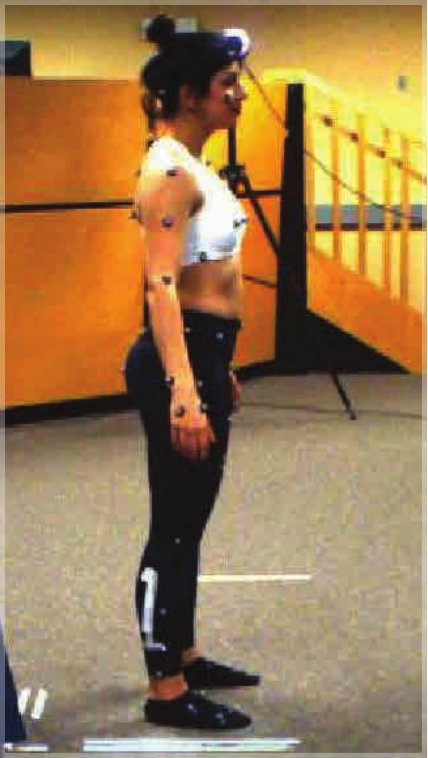}
    \label{fig:sfig1}
  \end{subfigure}%
  \begin{subfigure}{0.25\textwidth}
    \centering
    \includegraphics[width=0.9\textwidth]{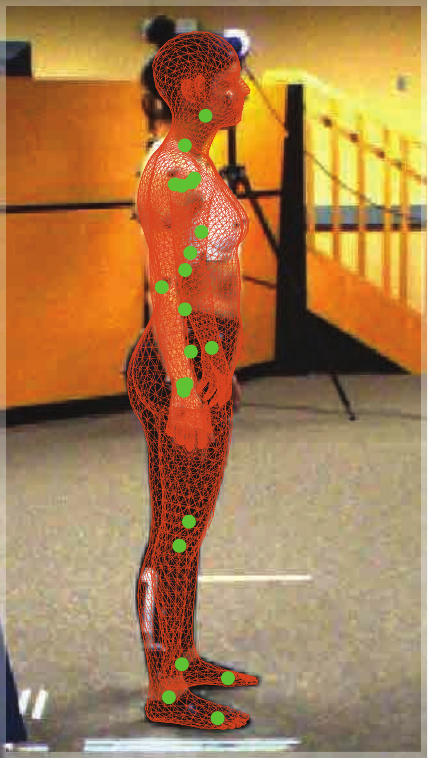}
    \label{fig:sfig1}
  \end{subfigure}%
\end{subfigure}

\begin{subfigure}{\linewidth}
  \begin{subfigure}{0.25\textwidth}
    \centering
    \includegraphics[width=0.9\textwidth]{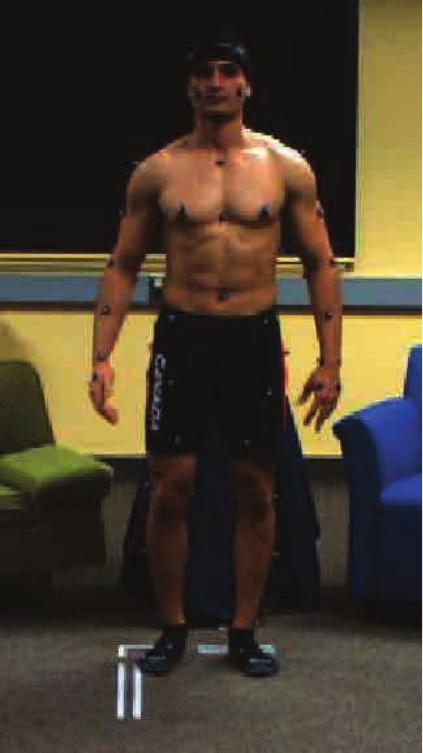}
    \label{fig:sfig1}
  \end{subfigure}%
  \begin{subfigure}{0.25\textwidth}
    \centering
    \includegraphics[width=0.9\textwidth]{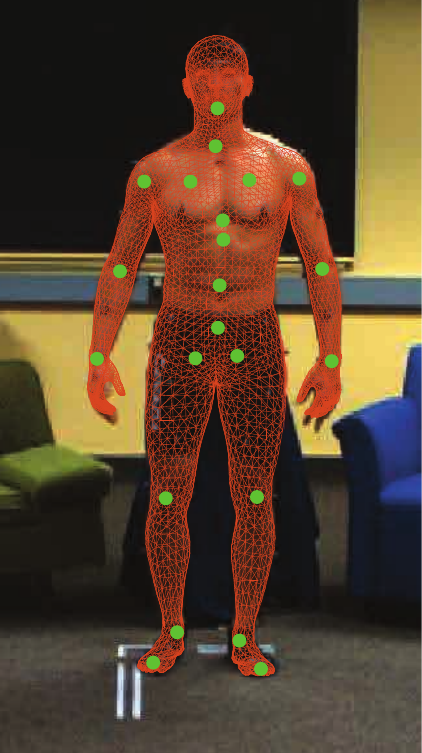}
    \label{fig:sfig1}
  \end{subfigure}%
  \begin{subfigure}{0.25\textwidth}
    \centering
    \includegraphics[width=0.9\textwidth]{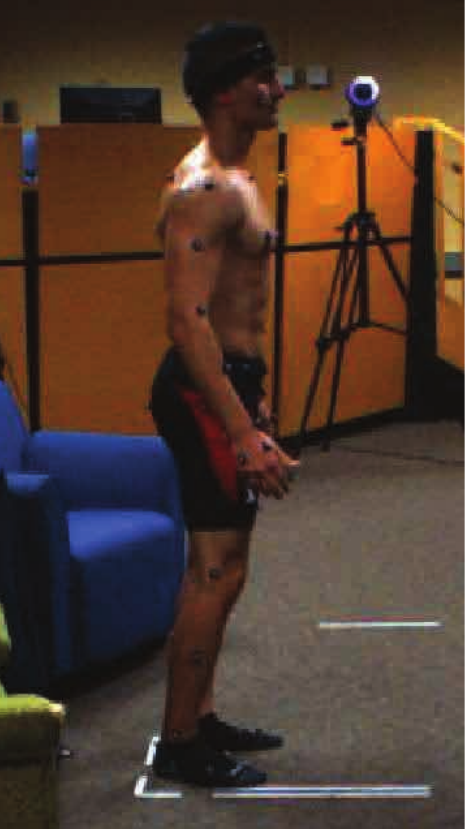}
    \label{fig:sfig1}
  \end{subfigure}%
  \begin{subfigure}{0.25\textwidth}
    \centering
    \includegraphics[width=0.9\textwidth]{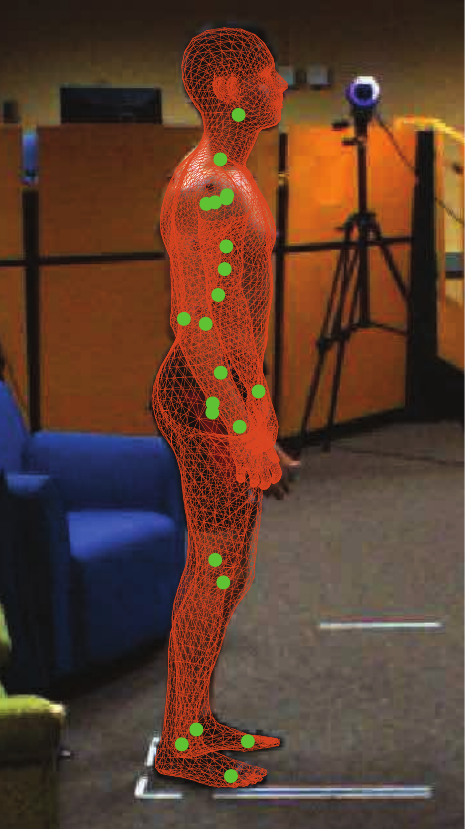}
    \label{fig:sfig1}
  \end{subfigure}
\end{subfigure}
\caption{Front and side view of aligned video frame, joint locations, and estimated body mesh (computed by MoSh++) for one female and male actor.}
\label{fig:samples}
\end{figure*}
\section{Introduction}
Recent advances in computer vision, in particular deep learning systems, have generated much interest in 2D human pose estimation~\cite{cao2018, toshev2014,pishchulin2016,fang2017,he2017,jain2013,tompson2014,sun2019,tang2018,ke2018}, 3D pose estimation~\cite{mehta2018,pavllo2019, varol2018,rhodin2018,neverova2018, sun2018,dabral2018,omran2018,alp2018,pavlakos2018}, human motion modelling~\cite{martinez2017, fragkiadaki2015,jain2016,holden2016,pavllo2018,holden2017,pavllo2019-2,ghosh2017,huang2019,barsoum2018}, 3D body reconstruction~\cite{kolotouros2019,pavlakos2018,pavlakos2019,kanazawa2018}, and activity recognition~\cite{simonyan2014,feichtenhofer2016, wang2016,carreira2017, varol2017,sharma2015,bilen2016,wang2015,liu2016,du2015,feichtenhofer2016-2} that are based on video data.
Large high-quality datasets are the cornerstone of such data-driven approaches. 

While there are many publicly available datasets of human motion recordings~\cite{sigal2010humaneva, mandery2015, de2009,trumble2017,ionescu2013}, they are limited in that they either contain data of a small number of different actors, use single hardware systems for motion recording, or provide unsynchronized data across different hardware systems. We overcome these limitations with our large Motion and Video dataset (MoVi) that contains five different subsets of synchronised and calibrated video, optical motion capture (MoCap), and inertial measurement units (IMU) data of 90 female and male actors performing a set of 20 predefined everyday actions and sports movements, and one self-chosen movement. MoVi is a multi-purpose human video and motion dataset designed for a variety of challenges such as human pose estimation, action recognition, motion modelling, gait analysis, and body shape reconstruction. To our knowledge, this is one of the largest datasets in terms of the recorded number of actors and performed actions.
 
The 3D ground truth skeletal pose in MoVi was computed using two different pipelines: V3D (bio-mechanics formulation)~\cite{C-Motion} and MoSh++ (regression model)~\cite{mahmood2019amass}. This allows a comparison of these two formulations and provides more options for the computed pose, depending on the tasks and challenges at hand (see section \ref{sec:Skeleton and Body Shape Extraction}).
MoVi is also part of the Archive of Motion Capture as Surface Shapes (AMASS)~\cite{mahmood2019amass}, available at \url{https://amass.is.tue.mpg.de/}. The approach of AMASS allows to estimate accurate body shape that is factorized into individual, pose-independent shape components and pose-dependent components for every single frame of the MoCap recordings. The resulting animated 3D meshes can be aligned with the camera coordinate system and be treated as ground truth 3D body shapes~(Figure \ref{fig:samples}).
\begin{figure}[h]
  \begin{center}
  \includegraphics[width=1\linewidth]{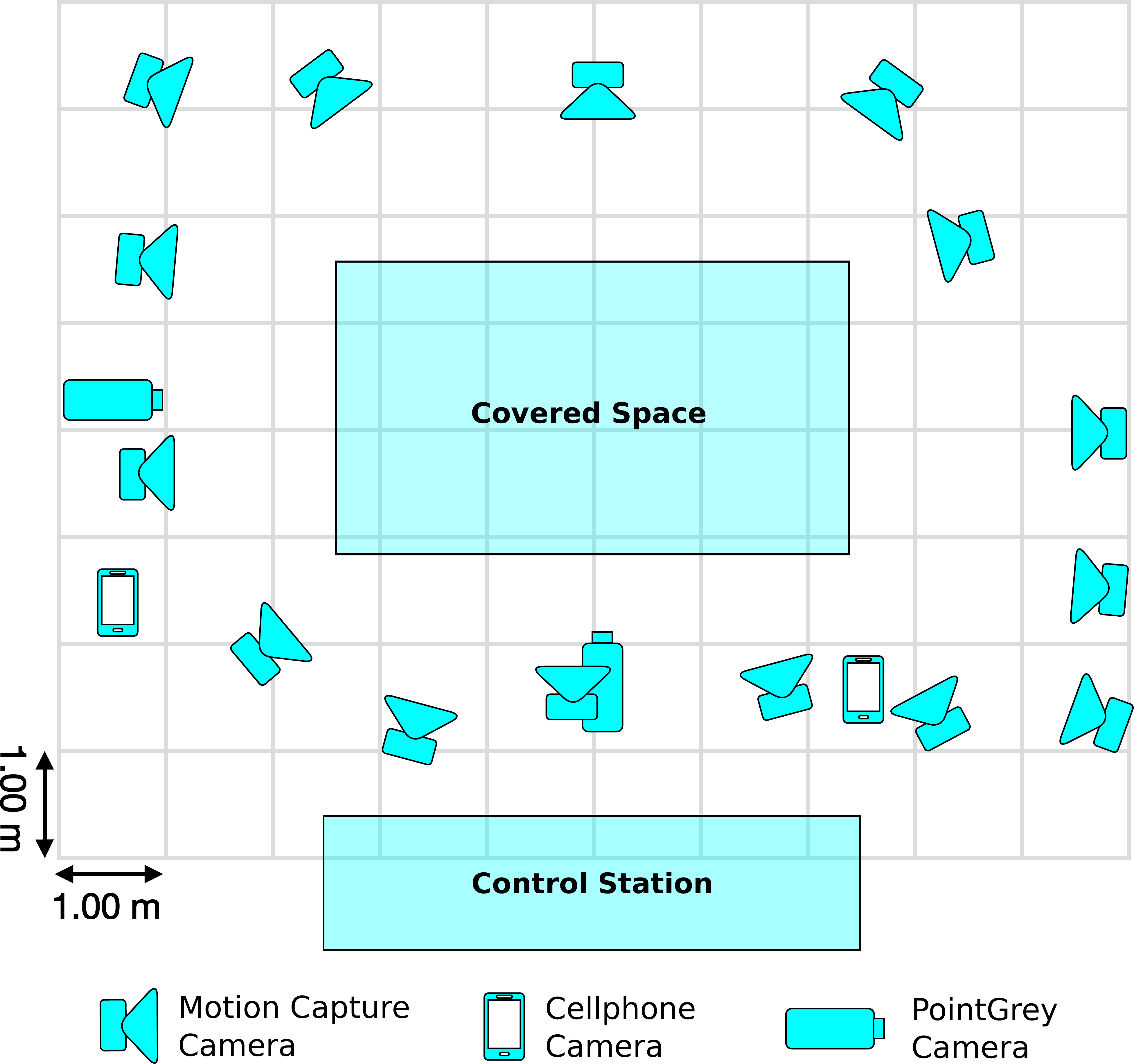}
  \end{center}
  \caption{Top view sketch of the capture room set-up. The position of the video cameras and motion capture cameras were arranged to cover a space of approximately 3 by 5 meters.}
  \label{fig:FloorPlan}
\end{figure}
\section{The MoVi Dataset} \label{section:MoViDataset}
\subsection{Summary of the Data} \label{section:Actors}
MoVi contains data from 90 subjects performing the same predefined set of 20 actions and one self-chosen movement in five rounds of data capturing. 90 people (60 females, 30 males; 5 left handed) were recruited from the local Kingston community. Descriptive statistics of all participants are shown in Table~\ref{tab:ParticipantCharacteristics}. Participants provided written informed consent. The experimental procedure was approved by the ethics committee of Queen's University, Kingston, and was performed in accordance with the Declaration of Helsinki.
\begin{table}[ht]
\centering
\begin{tabular}{p{3cm}p{2cm}p{2cm}} \\ \toprule
& Females & Males\\ \midrule\midrule
Height (m) & 1.65 (0.08) & 1.75 (0.06)\\ 
Weight (kg) & 60.35 (8.03) & 72.3 (10.98)\\ 
BMI (kg/m$^2$) & 22.16 (3.02) & 23.6 (3.24)\\
Age (y) & 20.47 (3.59) & 23.6 (3.61) \\ \bottomrule
\end{tabular}
\caption{Descriptive statistics (Mean (SD)) of the 60 female and 30 male actors.}
\label{tab:ParticipantCharacteristics}
\end{table}

The actors performed the same predefined set of 20 movements in a randomised order in five data capturing sequences. The movements included everyday actions and sports movements: (1) Walking, (2) Jogging, (3) Running in place, (4) Side gallop, (5) Crawling, (6) Vertical jumping, (7) Jumping jacks, (8) Kicking, (9) Stretching, (10) Cross arms, (11) Sitting down on a chair, (11) Crossing legs while sitting, (13) Pointing, (14) Clapping hands, (15) Scratching one's head, (16) Throwing and catching, (17) Waving, (18) Pretending to take a picture, (19) Pretending to talk on the phone, (20) Pretending to check one's watch. In each of the five sequences, the actors additionally performed one self-chosen motion (21).

The five sequences of data capturing differed in the hardware systems used to capture the motions, in participants' clothing (minimal, or normal), and whether or not there was a rest pose between successive motions. An overview of the different capture rounds is provided in Table~\ref{tab:CaptureRounds}, technical details of the hardware systems are provided in Table~\ref{tab:Hardware}.

Data capture sequence \textit{\textquotedblleft F\textquotedblright} was captured using the $67$ MoCap marker layout suggested in MoSh \cite{loper2014mosh}. Actors wore tight-fitting minimal clothing in order to minimize marker movement relative to the body. The markers were attached onto the actors' skin and clothes using double-sided tape. The MoCap system was synchronized with two video cameras capturing the actions from different viewpoints (front and side). Those two cameras were calibrated by computing the translation and rotation of the cameras relative to the coordinate system of the MoCap system. Two hand-held cellphone cameras were additionally used, however, the recordings were neither synchronized nor calibrated against the MoCap system. The different actions were separated by a rest A-pose. In our dataset, we provide the unedited full sequence of all motions, as well as trimmed MoCap and video sequences of the single motions. Our motivation for this capture round was to obtain accurate full skeletal (pose) information and frame-by-frame body shape parameters without any artifacts imposed by clothing. Therefore, this round can be considered more suitable for 2D or 3D pose estimation and tracking, and 3D shape reconstruction. The data collected in \textit{\textquotedblleft F\textquotedblright} was processed using two different pipelines: MoSh++~\cite{mahmood2019amass} and V3D~\cite{C-Motion} (see \ref{sec:Skeleton and Body Shape Extraction}). Example images of a female and male actor in rest pose are shown in Figure~\ref{fig:samples}.

To achieve more natural looking capture data, we recorded four more capture rounds where the actors wore normal clothing. Data capture rounds \textit{\textquotedblleft S1\textquotedblright} and \textit{\textquotedblleft S2\textquotedblright} were captured with a sparse set of $12$ MoCap markers (4 markers placed on the head, 2 on each ankle and 2 on each wrist) which allowed the participants to wear normal clothing. Having the attached markers we could accurately extract the main end-effectors including the head, wrists, and ankles. It further allowed us to synchronize the IMU data with the MoCap and video capture system (Section~\ref{section:Synchronization}). The actions were additionally recorded using synchronized computer vision cameras, cellphone cameras, and an IMU system. Whereas a rest A-pose separated the actions in \textit{\textquotedblleft S1\textquotedblright}, there was a natural transition between the different actions in \textit{\textquotedblleft S2\textquotedblright}. This setup was used as it allows to infer the pose and body shape by fusing a sparse marker set and IMU recordings while keeping the clothing natural.

\begin{figure*}
\centering
\begin{subfigure}{\linewidth}
    \begin{subfigure}{0.25\textwidth}
    \centering
    \includegraphics[width=0.9\textwidth]{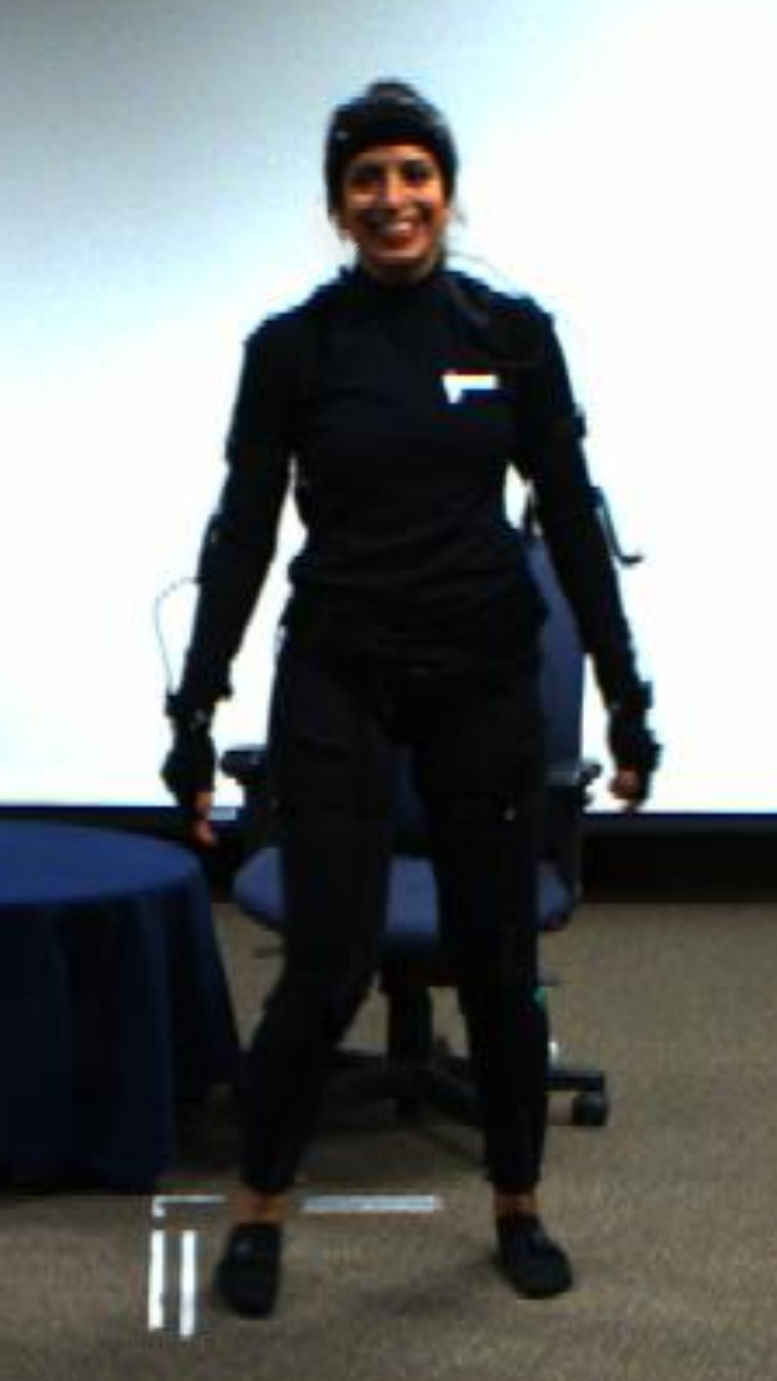}
    \end{subfigure}%
\begin{subfigure}{0.25\textwidth}
    \centering
    \includegraphics[width=0.9\textwidth]{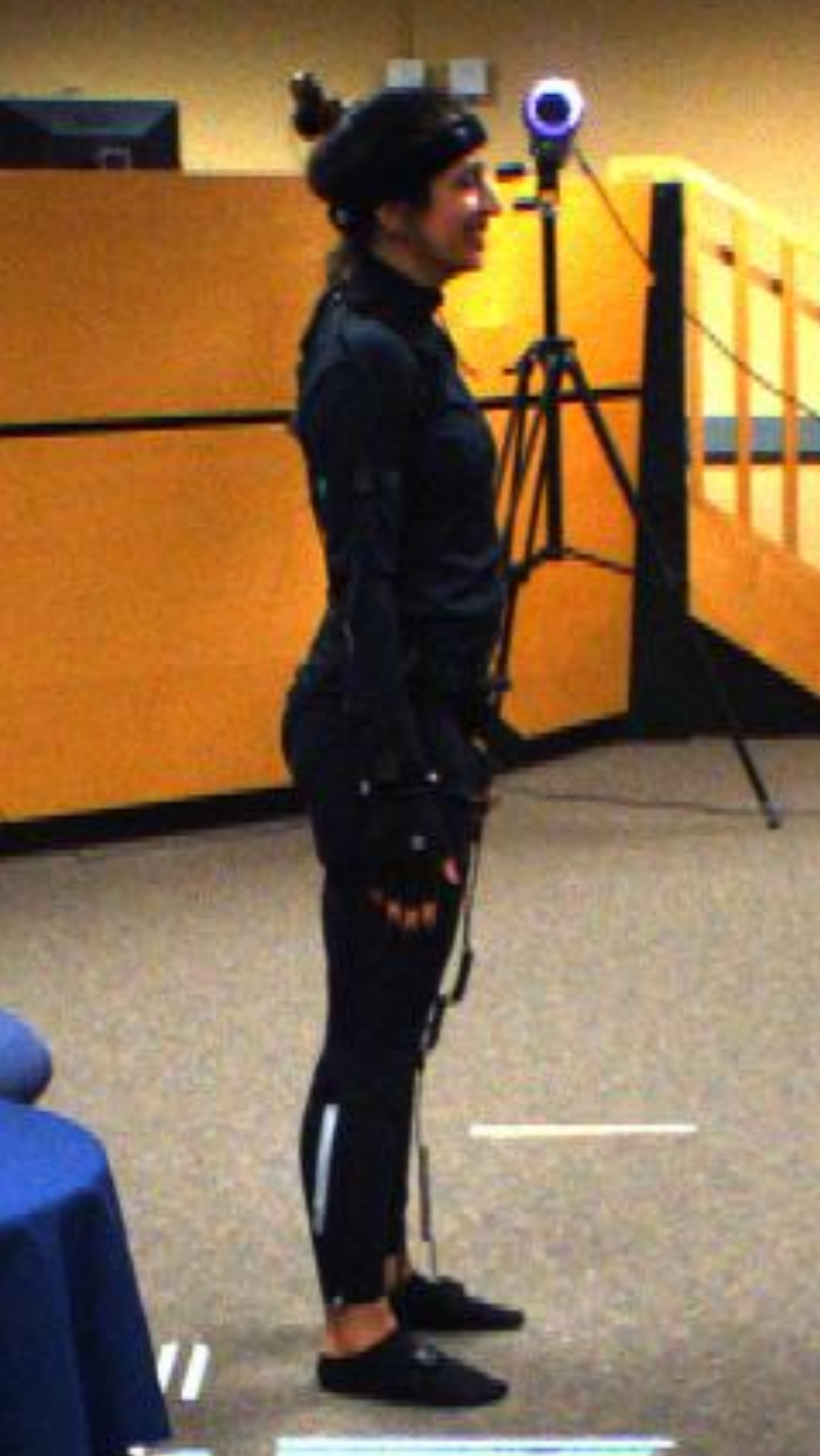}
  \end{subfigure}%
  \begin{subfigure}{0.25\textwidth}
    \centering
    \includegraphics[width=0.9\textwidth]{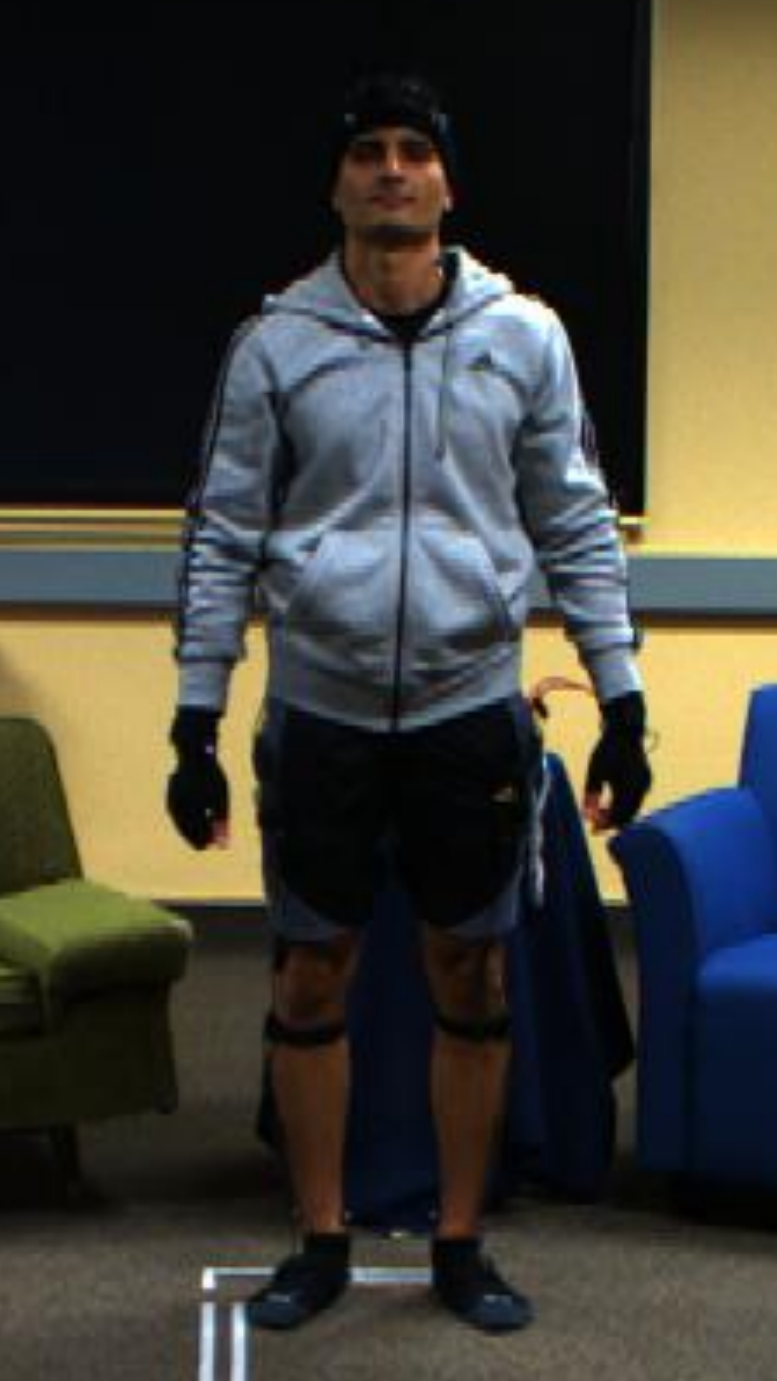}
  \end{subfigure}%
  \begin{subfigure}{0.25\textwidth}
    \centering
    \includegraphics[width=0.9\textwidth]{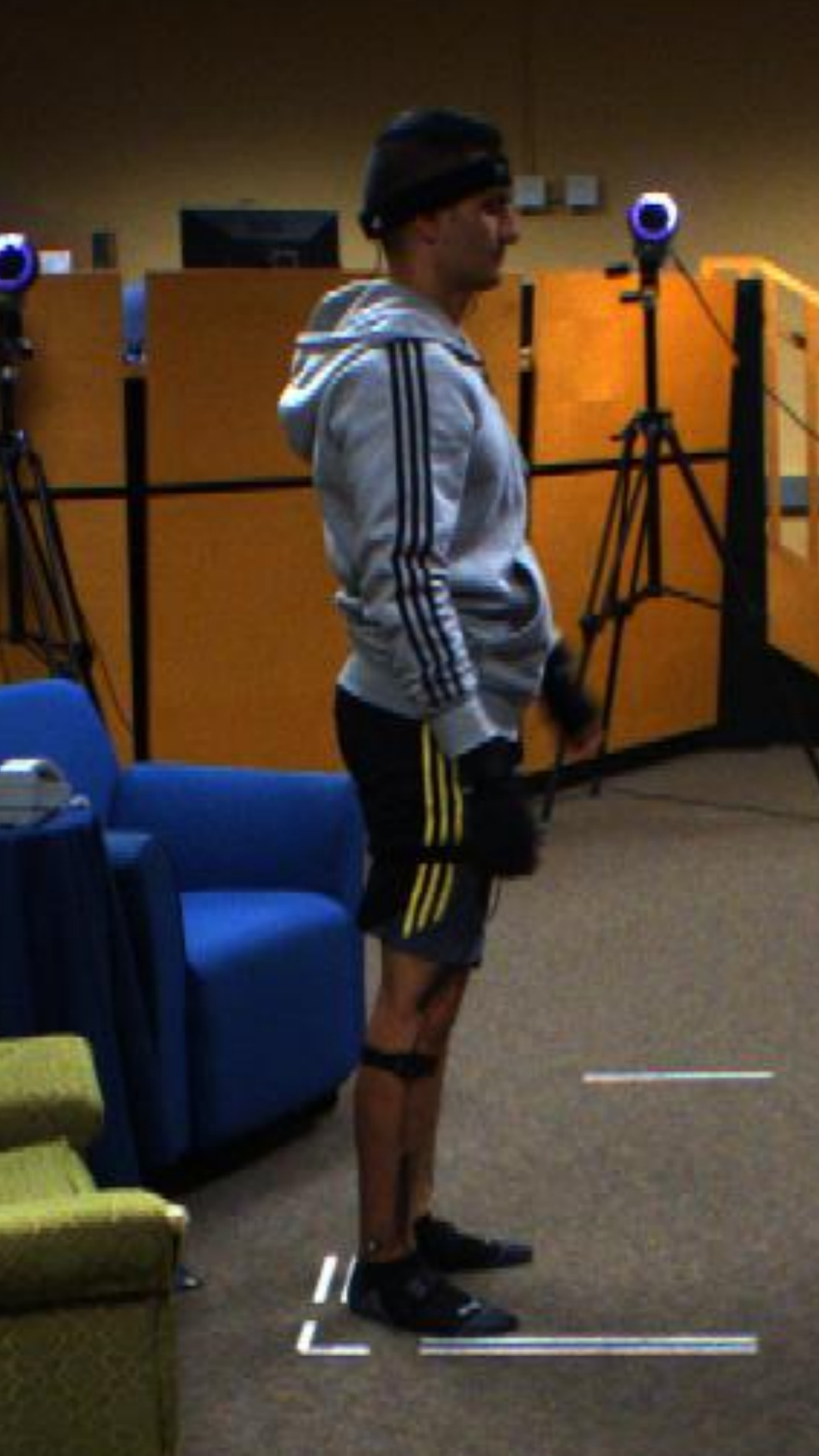}
  \end{subfigure}%
\end{subfigure}
\caption{Example pictures of one female and one male actor wearing the IMU suits used for the capture rounds S1, S2, I1, and I2.}
\label{fig:samples2}
\end{figure*}

While real clothing is essential for many meaningful data, it precludes the use of certain motion capture techniques. The data capture rounds \textit{\textquotedblleft I1\textquotedblright} and \textit{\textquotedblleft I2\textquotedblright} were thus captured using only IMU and video cameras (not synchronised). Motions in \textit{\textquotedblleft I1\textquotedblright} are separated by a rest A-pose, whereas there is a natural transition between the different actions in \textit{\textquotedblleft I2\textquotedblright}. The data collected in \textit{\textquotedblleft I1\textquotedblright} and \textit{\textquotedblleft I2\textquotedblright} is suitable for researchers that aim for computing pose or body shape without any artifacts imposed by the optical markers. The examples of the IMU suit used for \textit{\textquotedblleft S1\textquotedblright}, \textit{\textquotedblleft S2\textquotedblright}, \textit{\textquotedblleft I1\textquotedblright}, and \textit{\textquotedblleft I2\textquotedblright} are shown in Figure~\ref{fig:samples2}. These recordings thus promise to enable a broad range of real-world applications.

\begin{table*}[ht]
  \centering
  \small
  \begin{tabular}{p{3.5cm}p{2.4cm}p{2.4cm}p{2.4cm}p{2.4cm}p{2.4cm}} \toprule
  Round & F & S1 & S2 & I1 & I2 \\ \midrule\midrule
  MoCap marker set & 67 & 12 & 12 & -- & -- \\  \midrule
  Video capture & yes & yes & yes & yes & yes \\  \midrule
  IMU & no & yes & yes & yes & yes \\  \midrule
  A-pose between motions & yes & yes & no & yes & no \\  \midrule
  Actor clothing & minimal & normal clothing & normal clothing & normal clothing & normal clothing \\  \midrule
  Length (min per person) & $\thicksim 2.7$ & $\thicksim 2.7$ & $\thicksim 1.7$ & $\thicksim 2.7$ & $\thicksim 1.7$  \\\bottomrule
  \end{tabular}
  \caption{Overview of the five difference capture rounds. F = Full; S = sparse marker set + IMU; I = IMU.}
  \label{tab:CaptureRounds}
\end{table*}

\subsection{Hardware} \label{section:Hardware}
The movements were captured using two different hardware systems, an optical motion capture system and an inertial measurement unit system. We used a commercial optical motion capture system from Qualisys with fifteen 1.3 MP cameras that provide the 3D location of passive reflective markers with a frame rate of 120 per second. For the IMU system, we used the Noiton Neuron Edition V2 which comes as a suit attached with 18 IMU sensor. Each sensor is composed of a 3-axis gyroscope, a 3-axis accelerometer, and a 3-axis magnetometer working with 120 fps. In addition to the global acceleration data, the IMU suit provides 3D displacements, speed, quaternions, and rotational speed for each joint. For distinct questions different hardware will be useful. For example, IMUs are great at acceleration, while other capture systems produce high precision. The use of different, complementary hardware system promises to allow a broad range of meaningful analyses. 

Video data was collected using two different types of cameras, smartphone cameras and computer vision cameras. We used two hand-held IPhone 7 smartphone cameras with a \({800 \times 600}\) resolution, global shutter, and 30 fps. As opposed to the computer vision cameras, the footage obtained with those smartphone cameras is shaky due to natural arm and hand movements. The video quality is similar to what the majority of commercially available smartphone cameras provides to date.
For the two computer vision cameras we used Grasshopers cameras from FLIR Inc company with \({800 \times 600}\) Sony ICX285 CCD sensors. The recordings of the FLIR cameras are synchronized with the MoCap cameras with 30 fps (aligned with every forth frame of the MoCap system). Detailed information of the used hardware is provided in Table \ref{tab:Hardware}. Figure \ref{fig:FloorPlan} shows the top-view floor plan and the location of the MoCap and video cameras. The process of how the devices were synchronized are described in Section~\ref{section:Synchronization}, the camera calibration is describes in Section~\ref{section:Calibration}.

\begin{table*}[ht]
\centering
\small
\begin{tabular}{p{4cm}p{4cm}p{7cm}} \\ \toprule
\multirow{4}{*}{MoCap System} & Brand and model & Qualisys Oqus 300 and 310 \\
& Number of cameras & 15 \\ 
& Resolution & 1.3 megapixel \\
& Frame rate & 120 Hz \\ 
& Synchronized & Yes \\\midrule
\multirow{7}{*}{Video Capture Systems 1} & Brand and model &  FLIR, Grasshopper\\
& Number of cameras & 2 \\
& Cameras synchronized & Yes \\
& Cameras calibrated & Yes \\
& Resolution & 800 x 600 pixels, 72 dpi, 24 bit depth \\
& Sensor & Sony ICX285, CCD \\ 
& Frame rate & 30 Hz \\ 
& File type & JPEG frames and AVI video files \\ \midrule
\multirow{7}{*}{Video Capture Systems 2} & Brand and model &  IPhone 7 rear camera \\
& Number of cameras & 2 \\
& Cameras synchronized & No \\
& Cameras calibrated & No \\
& Resolution & 1920 x 1080 pixels\\
& Sensor & Sony Exmor RS, CMOS \\ 
& Frame rate & 30 Hz \\ 
& File type & MP4 video files \\ \midrule
\multirow{8}{*}{IMU} & Brand and model  &  Noitom, Neuron Edition V2\\
& Number of sensors & 18 Neurons \\ 
& Sensor & 9-axis IMU composed of 3-axis gyroscope, 3-axis accelerometer, and 3-axis magnetometer \\
& Synchronized & in S1 and S2 rounds \\
& Frame rate & 120 Hz \\
& File type & BVH and calculation files\\ \bottomrule
\end{tabular}
\caption{Technical details of the hardware systems used to capture the MoVi dataset.}
\label{tab:Hardware}
\end{table*}

\subsection{Synchronization}\label{section:Synchronization}
\subsubsection*{MoCap and Video}\label{section:MoCap and Video}
To provide a frame-by-frame accurate 3D motion overlaid on the video footage, the motion capture system should be synchronized with cameras in frame and then calibrated to the same coordinate.
The synchronization between motion capture cameras and the FLIR Grasshopers video cameras was done in hardware. In our setup, the video cameras where triggered by the synchronization signal provided by the MoCap system. Due to the frame rate limits in video cameras, the synchronization frequency was divided by 4 which reduced the video capture frame rate to 30 fps. The phone cameras were not synchronized with the motion capture cameras.

\subsubsection*{IMU}\label{section:IMU}
In round \textit{\textquotedblleft S1\textquotedblright} and \textit{\textquotedblleft S2\textquotedblright}, we used a reduced optical motion marker set layout with $12$ markers. Although the main motivation for using this reduced marker set was to allow the actors to wear natural clothing, this small set of markers offers several advantages: 1) It provides sparse but accurate data for some of the main joints (head, wrists, and ankles). The data can be applied in a data fusion approach along with IMU data to infer the exact joint locations. We leave this to future work. 
2) It allowed us to synchronized IMU and MoCap data. To synchronize the data, we used cross-correlation between these two modalities. The two coordinate systems were not aligned, however, the differences between the orientation of the two \(z\) axes is negligible: the \(z\) axis of the IMU coordinate system is oriented towards gravity, while \(z\) axis in MoCap coordinate system is perpendicular to the floor. Because the MoCap system was synchronised with the video cameras (Section~\ref{section:MoCap and Video}), we additionally obtained synchronized IMU and video data.

Suppose \({v}_{z}^{j}(t) \) and \(\tilde{v}_{z}^{j}(t) \) are the \(z\) component of tracked position of joint \(j\) recovered by the motion capture and IMU systems, respectively (we are using the 3D positions provided by the IMU software instead of double-integrating over accelerations).
The synchronization parameters, temporal scale \(\alpha\) and temporal shift \(\beta\), are found by maximizing:

\begin{eqnarray}
  \label{eq:IMU_synchronization}
    \max_{\alpha, \beta}  \max_{\tau}({v}_{z}^{j}(t)\star \tilde{v}_{z}^{j}(\alpha t + \beta))(\tau),
  \end{eqnarray}
where
\begin{eqnarray}
  \label{eq:xcorr}
  ({v}_{z}^{j}(t)\star \tilde{v}_{z}^{j}(\alpha t + \beta))(\tau) \triangleq \int_{-\infty}^{\infty} {v}_{z}^{j}(t) \tilde{v}_{z}^{j}(\alpha t + \beta + \tau) d t,
\end{eqnarray}

is the cross-correlation between \({v}_{z}^{j}(t) \) and shifted-and-scaled version of \(\tilde{v}_{z}^{j}(t) \). $\alpha$ and $\beta$ are the scale and shift parameters, respectively. The optimal parameters are those which achieve the highest peak in cross-correlation.

The procedure mentioned above was done for left and right ankles and checked qualitatively for all data.

\begin{table*}[h]
  \small
  \begin{center}
  \begin{tabular}{p{5cm} p{1cm} p{9cm}}
    \texttt{F\_amass\_subject\_$\langle$ID$\rangle$.mat}& \centering-- &Contains the full marker set MoCap data processed by MoSh++ in the AMASS project and augmented with 3D joints' positions and metadata. All files are compressed and stored as 
    \texttt{F\_AMASS.rar}. The original \texttt{npz} files and the rendered animation files are available at \url{https://amass.is.tue.mpg.de/}\\ \\
    \texttt{$\langle$round$\rangle$\_v3d\_subject\_$\langle$ID$\rangle$.mat}& \centering-- &Contains the MoCap data processed by V3D and augmented with metadata. All files are compressed as \texttt{subject\_1\_45\_F\_V3D.rar} which contains \textit{\textquotedblleft F\textquotedblright} round data from subject 1 to 45, \texttt{subject\_46\_90\_F\_V3D.rar} which contains \textit{\textquotedblleft F\textquotedblright} round data from subject 46 to 90, and \texttt{S\_V3D.rar}  which contains \textit{\textquotedblleft S1\textquotedblright} and \textit{\textquotedblleft S2\textquotedblright} rounds data from all subjects.\\\\
    \texttt{imu\_subject\_$\langle$ID$\rangle$.mat}& \centering-- &Contains the processed IMU \texttt{calculation} files augmented with metadata. Each file contains the data collected in all \textit{\textquotedblleft S1\textquotedblright}, \textit{\textquotedblleft S2\textquotedblright}, \textit{\textquotedblleft I1\textquotedblright}, \textit{\textquotedblleft I2\textquotedblright} rounds.  All files are compressed as \texttt{IMU\_calc.rar} \\ \\
    \texttt{$\langle$round$\rangle$\_imu\_subject\_$\langle$ID$\rangle$.bvh}& \centering-- & Contains the \texttt{bvh} files generated by the IMU software. All files are compressed as \texttt{IMU\_bvh.rar} \\ \\
    \texttt{$\langle$round$\rangle$\_$\langle$camera$\rangle$\_subject\_$\langle$ID$\rangle$.avi}& \centering-- &An AVI video data for each subject, round, and camera\\ \\
    \texttt{cameraParams\_$\langle$camera$\rangle$.mat}& \centering-- &Contains the camera calibration data. Each file contains the MATLAB intrinsic camera parameter object\\ \\
    \texttt{Extrinsics\_$\langle$camera$\rangle$.mat}& \centering-- &Contains the camera extrinsics parameters (rotation matrix and translation vector)\\ \\
  \end{tabular}
\end{center}
\caption{MoVi dataset file stucture. \texttt{$\langle$ID$\rangle\in$ \{1,2,\dots,90\}} is the subject number. \texttt{$\langle$round$\rangle\in$ \{F,S1,S2,I1,I2\}} is the data collection round (see Table \ref{tab:ParticipantCharacteristics}). \texttt{$\langle$camera$\rangle\in$ \{PG1,PG2,CP1,CP2\}} is the camera type, where PG and CP indicate the computer vision and cellphone cameras, respectively}
\label{tab:naming}
\end{table*}
\subsection{Calibration}\label{section:Calibration}
The calibration of the MoCap cameras were done by a measurement procedure in Qualisys Track Manager software \cite{qualisys}. The software allows to compute the orientation and position of each camera in order to track and perform calculations on the 2D data into 3D data. 
To compute the computer vision cameras' intrinsics and lens distortion parameters, we used the MATLAB Single Camera Calibrator~\cite{heikkila1997four,zhang2000flexible,Bouguet}, where focal length ($F\in \mathbb{R}^{2} $), optical center ($C\in \mathbb{R}^{2} $), skew coefficient ($S\in \mathbb{R}$), and radial distortion ($D\in \mathbb{R}^{2} $) are estimated for each camera.\\
The extrinsic parameters which represent the rotation \(R \in SO(3)\) and translation \(T \in \mathbb{R}^{3}\) transformations from world coordinates (MoCap coordinate system) to camera coordinates, are estimated using the semi-automated method proposed by Sigal et al.~\cite{sigal2010humaneva}. The trajectory of a single moving marker was recorded by synchronized MoCap and video cameras for around \(2000\) synchronized frames. Given the recorded 3D positions of the marker in MoCap coordinates as world points and the 2D positions of the marker in the camera frame as image points, the problem of finding the best 2D projection can be formulated as a Perspective-\textit{n}-Point (PnP) problem where the Perspective-Three-Point (P3P) algorithm~\cite{gao2003complete} is used to minimize the re-projection error as follows:

\begin{eqnarray}
  \label{eq:xcorr}
  \min _{R, T}\sum_{t=1}^{N}\left\|P_{2D}-f\left(P_{3D}; R, T, K\right)\right\|^{2},
\end{eqnarray}
where $f$ is the projection function and $K\in \{F, C, S, D\}$ is the set of camera intrinsics parameters.

\subsection{Skeleton and Body Shape Extraction from MoCap data}\label{sec:Skeleton and Body Shape Extraction}
The skeleton (joint locations and bones) were computed with two different pipelines.
Visual 3D software (manufacturer C-MOTION): Visual 3D is an advanced biomechanics analysis software for 3D motion capture data~\cite{C-Motion}. In our V3D pipeline, pelvic segment was created using CODA~\cite{Coda} and the hip joints positions were estimated by Bell and Brand hip joint center regression \cite{bell1989,bell1990}. The upper body parts were estimated using Golem/Plug-in Gait Upper Extremity model as implemented in Vicon~\cite{C-Motion}. The skeleton is represented by 20 joints in two different formats: 1) in joint angles, that is the angle of each bone relative to coordinate system of its parent joint, and 2) as global 3D joint locations.

MoSh++: MoSh++\cite{mahmood2019amass} is an approach which estimates the body shape, pose, and soft tissue deformation directly from motion capture data. Body shape and pose are represented using a rigged body model called SMPL\cite{loper2015smpl} where the pose is defined by joint angles and shape is specified by shape blend shapes. MoSh++ uses a generative inference approach whereby the SMPL body shape and pose parameters are optimized to minimize reconstruction errors. The skeletal Joints location are computed using a linear regression function of mesh vertices. The estimated SMPL body is extended by adding dynamic blend shapes using the dynamic shape space of DMPL. Each frame in the "MoShed" representation includes 16 SMPL shape coefficients, 8 DMPL dynamic soft-tissue coefficients, and 66 SMPL pose coefficients as joint angles (21 joints + 1 root). MoShed data was computed in collaboration with the authors of AMASS~\cite{mahmood2019amass}.

The main difference between the skeleton represented by MoSh and the skeleton represented by V3D is that the MoShed version is generally more robust to occlusion because it uses distributed information, rather than doing the computations locally. This makes it a better choice for the task of pose estimation and tracking as the joints locations are all available over the time. However, the estimated joint location can be noisy during the occlusion and the error may propagate to other joints too. On the other hand, V3D provides a more accurate estimation of joint location. Therefore, one may prefer using the V3D joints representation for the task of gait analysis. The only drawback of V3D representation compared to MoSh++ is that the joints cannot be computed when a related marker is occluded.

Our dataset is the first sizable dataset including not only 3D joint locations, but also a highly accurate 3D mesh of the body which can be projected onto the video recordings. This can be useful for approaches that try to estimate body shape from video data.
\section{Dataset structure}
We used the Dataverse repository to store the motion and video data. We provide the original \texttt{AVI} video files to avoid any artifacts added by compression methods. The processed MoCap data is provided in two different versions based on the post-processing pipeline (AMASS and V3D). We provide joint angles and joint 3D locations computed by both pipeline along with the associated kinematic tree, occlusions, and optical marker data. Synchronized IMU data (along with the original data) are computed by processing \texttt{calculation} files (see Section~\ref{section:IMU}) and converted to \texttt{mat} format which provides raw acceleration data, displacement, velocity, quaternions and angular velocity. The \texttt{bvh} files generated by the IMU software are also provided on the website. The support code to use data in MATLAB and Python environments is also provided. The dataset naming structure is provided in Table \ref{tab:naming}.

\section{Discussion and Conclusion}
We provide the large Motion and Video dataset MoVi, which is now available at \url{https://www.biomotionlab.ca/movi}. The dataset contains motion recordings (optical motion capture, video, and IMU) of 90 male and female actors performing a set of 20 everyday actions and sports motions, and one additional self-chosen motion. The different sequences of the dataset contain synchronized recordings of the three different hardware systems. In addition, our full-body motion capture recordings are available as realistic 3D human meshed represented by a rigged body model as part of the AMASS dataset \cite{mahmood2019amass}. This allows video overlay of not only the body joints, but also the full body meshes. To our knowledge, MoVi is the first dataset with synchronized pose, pose-dependent shape, and video recordings. The multi-modality makes our dataset suitable for a wide range of challenges such as human pose estimation and tracking, body shape estimation, human motion prediction and synthesis, action recognition, and gait analysis.

\section*{Acknowledgments}
We wish to thank Nima Ghorbani for post-processing our motion capture data so it could be added to the AMASS dataset (https://amass.is.tue.mpg.de/), and all others authors of AMASS for their approval to add the processed data to our dataset. We further wish to thank Viswaijt Kumar for his help with post-processing the data and setting up the data repository and website. This research was funded by a NSERC Discovery Grant and contributions from CFREF VISTA to NFT.

{\small
\bibliographystyle{ieee_fullname}
\bibliography{MoVi.bbl}
}

\end{document}